\documentclass[a4paper, 10pt, conference]{ieeeconf}      
\usepackage{FG2020}

\FGfinalcopy 

\IEEEoverridecommandlockouts                              
\usepackage{graphics} 
\usepackage{epsfig} 
\usepackage{mathptmx} 
\usepackage{times} 
\usepackage{amsmath} 
\usepackage{amssymb}  
\usepackage{multirow}

\def\FGPaperID{166} 

\title{\LARGE \bf First Investigation Into the Use of Deep Learning for Continuous Assessment of Neonatal Postoperative Pain}

\author{\parbox{16cm}{\centering
    {\large Md Sirajus Salekin$^1$, Ghada Zamzmi$^{1}$, Dmitry Goldgof$^{1}$, Rangachar Kasturi$^{1}$, Thao Ho$^{2}$ and Yu Sun$^{1}$}\\
    {\normalsize
    $^1$ Department of Computer Science and Engineering, University of South Florida, FL, USA\\
    $^2$ College of Medicine Pediatrics, USF Health, University of South Florida, FL, USA}}
    \thanks{Partial support through UNI: \lq\lq Dmitry Goldgof acknowledges partial support through a University of South Florida Nexus Initiative (UNI) Award.\rq\rq}
}

\begin{document}

\onecolumn
\noindent \huge IEEE Copyright Notice\\

\noindent \normalsize \textcopyright \hspace{0.05cm} 2020 IEEE. Personal use of this material is permitted. Permission from IEEE must be obtained for all other uses, in any current or future media, including reprinting/republishing this material for advertising or promotional purposes, creating new collective works, for resale or redistribution to servers or lists, or reuse of any copyrighted component of this work in other works.\\

\noindent \large {Accepted to be Published in: Proceedings of the 2020 IEEE International Conference on Automatic Face and Gesture Recognition (FG 2020), 2020.}
\twocolumn
\normalsize 

\ifFGfinal
\thispagestyle{empty}
\pagestyle{empty}
\else
\author{Anonymous FG2020 submission\\ Paper ID \FGPaperID \\}
\pagestyle{plain}
\fi
\maketitle

\begin{abstract}
This paper presents the first investigation into the use of fully automated deep learning framework for assessing neonatal postoperative pain. It specifically investigates the use of Bilinear Convolutional Neural Network (B-CNN) to extract facial features during different levels of postoperative pain followed by modeling the temporal pattern using Recurrent Neural Network (RNN). Although acute and postoperative pain have some common characteristics (e.g., visual action units), postoperative pain has a different dynamic, and it evolves in a unique pattern over time. Our experimental results indicate a clear difference between the pattern of acute and postoperative pain. They also suggest the efficiency of using a combination of bilinear CNN with RNN model for the continuous assessment of postoperative pain intensity. 
\end{abstract}


\section{Introduction}
Postoperative pain, or pain after surgery, occurs as a result of a tissue injury and it usually lasts for up to seven days \cite{gan2017poorly}. The inadequate treatment of postoperative pain leads to chronic pain and, therefore, increases the financial burden on the patients and significantly reduce the quality of their life. It can also lead to serious physiological outcomes \cite{gan2017poorly} such as changes in respiratory, cardiovascular, and immune functions \cite{gan2017poorly}. In addition to the physiological outcomes, inadequate management of postoperative pain can lead to impaired sleep, depression, and anxiety \cite{gan2017poorly}. Because high medication doses have several side effects (e.g., drowsiness and addiction\cite{gan2017poorly}), the main goal of postoperative pain assessment and management is to provide maximum pain relief with minimum side effects. 

Neonatal postoperative pain is currently assessed using validated multidimensional score-based scales such as N-PASS (Neonatal Pain, Agitation and Sedation Scale) \cite{hummel2008clinical} and PIPP (premature infant pain profile) \cite{stevens2014premature}. These scales integrate several physiological (e.g., vital signs) and behavioral (e.g., facial expression, body movement) indicators to assess the neonate's state as: normal, slight pain, moderate pain, severe pain, moderate sedation, or severe sedation. This score-based standard for assessing pain suffers from inconsistency and discontinuity. The inconsistency of assessment occurs as a result of the variations between observers due to different idiosyncratic factors (e.g., gender and experience). The discontinuity of assessment can lead to missing pain changes and patterns, and therefore, result in inaccurate management. Recently, health professionals emphasized, in a guideline for managing postoperative pain \cite{aubrun2019revision}, the importance of frequent assessment since pain is a temporal event that changes in a particular pattern over time. 

To mitigate these limitations, several traditional and automated solutions were proposed by clinicians and engineers in the past decade. Presentations of existing automated solutions or methods for assessing neonatal procedural pain are presented in \cite{zamzmi2018review}. In case of postoperative pain, Sikka et al. \cite{sikka2015automated} used traditional handcrafted (action unit-based) method for assessing postoperative pain in children. To the best of our knowledge, our work is the first to investigate the automated assessment and monitoring of neonatal postoperative pain using deep learning methods. Specifically, it introduces a method for monitoring and assessing the intensity of neonatal postoperative pain. It makes the following contributions:
\begin{itemize}
    \item It proposes a fully automated deep learning-based method for assessing the intensity of postoperative pain in neonates. 
    \item It presents the first investigation into the use of bilinear CNN \cite{lin2015bilinear} with LSTM \cite{hochreiter1997long} for pain recognition. The bilinear model, which consists of two CNNs, allows modeling local pairwise feature interactions in a translationally invariant manner while LSTM allows modeling temporal pain pattern.
    \item It presents a neonatal pain dataset, with over 600 minutes, collected from premature and newborns while they are undergoing postoperative pain. The dataset includes visual (face and body), vocal, and physiological signals. It also includes manual pain scores documented by trained NICU nurses. To the best of our knowledge, this is the first neonatal postoperative pain dataset.
\end{itemize}

Section II presents technical background. Section III describes our pain dataset. We present our methodology in Section IV followed by the experimental results in Section V. Finally, Section VI concludes the paper. 


\section{Technical Background}
This section provides brief introductions to the architectures that are used to develop the proposed framework. These architectures include VGG16, bilinear CNN, and LSTM. 

\subsection{VGG Architecture}
VGG \cite{simonyan2014very} is a state-of-the-art architecture that consists of several convolution blocks. Each block has several convolutional layers followed by a max-pooling layer. Finally, 3 consecutive Fully Connected (FC) layers are added as classification layers. VGG16 made significant improvement over AlexNet by using large kernel size filters. Although VGG has several architectures, VGG16 and VGG19 are the most popular since they achieved state-of-the-art accuracy in different classification tasks. In this paper, two pre-trained VGG16 networks, trained on VGGFace2 \cite{cao2018vggface2} and ImageNet \cite{deng2009imagenet} datasets, are used to build and train the bilinear model.

\subsection{Bilinear CNN}
Bilinear CNN \cite{lin2015bilinear} model is a recent novel solution for fine-grained image classification. It can effectively collect local pairwise information and produce orderless texture features. This Bilinear model merges two CNN models through a bilinear vector. Mathematically, a bilinear model $B$ can be represented as $B = (F_X, F_Y, P, C)$, where $F_X$ and $F_Y$ are the two feature functions applied on an image $I$ and a location $L$, $P$ is the pooling function, and $C$ is the classification function. The features of the two functions are then merged as follows.
\begin{equation}
 B = (I, L, F_X, F_Y) \xrightarrow{} F_X(I, L)^T F_Y(I, L)
\end{equation}
To create the feature descriptor, sum pooling is applied across the image to gather all the bilinear features. The final bilinear vector $u = \sum B{(I, L)}$ is passed to a signed square root and a $l_2$ normalization step as follows.
\begin{equation}
    v \xleftarrow[]{ sqrt}(sign(u)*\sqrt{\left|{u}\right|})
\end{equation}
\begin{equation}
    w \xleftarrow[]{normalization}(v/{||{v}||_2})
\end{equation}
As the entire network creates a directed acyclic graph, the network parameters can be easily updated through the backpropagation by end-to-end training. This type of model can robustly handle intra-class variations caused by large pose, lighting, and background \cite{lin2015bilinear} variations, which are common in our postoperative pain dataset.

\subsection{LSTM Network}
Recurrent Neural Network (RNN) is typically used for modeling temporal changes. RNN uses the past state to update the current and future states. RNN suffers from lack of preserving long-term dependencies \cite{bengio1994learning, hochreiter1997long}. Hence, Long Short-Term Memory (LSTM) \cite{hochreiter1997long} was introduced. LSTM solves the problem of long-term dependency by the controlled memory cell of input, output, and forget gate. In this work, we used the deep features extracted by the Bilinear CNN model to train the LSTM network to capture temporal changes in pain intensity.


\section{Neonatal Pain Assessment Dataset}
Our dataset contains acute and postoperative pain data (visual, vocal, and physiological) collected from 45 neonates during their NICU hospitalization at Tampa General hospital. Portion of this dataset is available at no cost to other academic researchers after signing an agreement\footnote{{https://rpal.cse.usf.edu/project\textunderscore neonatal\textunderscore pain/}} form and obtaining a clearance. 

\subsection{Equipment Setup}
\begin{figure}[tb]
\centering
\includegraphics[width=.40\textwidth]{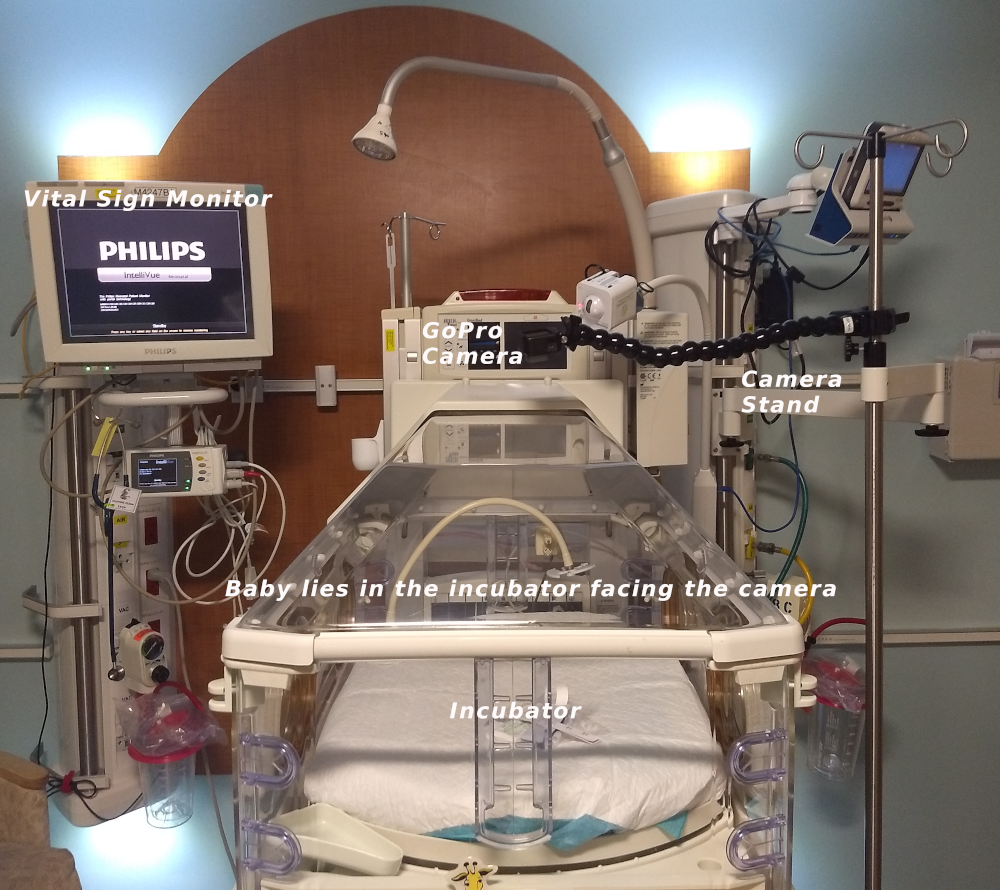}
\caption{Equipment setup of the data collection}
\label{fig_equipment_setup}
\end{figure}

A GoPro Hero 5 Black camera was used to record audio and video signals from the neonates. The recording resolution is 1080p with a frame rate of 30 FPS. The camera was installed in a camera stand and positioned facing the incubator to capture the face and body regions. Most of the neonates were lying supine facing the camera; some neonates were lying prone. In addition to the vocal and visual data, physiological data (heart rate, breathing pattern, saturation rate, and blood Pressure) were also simultaneously collected from the bedside vital sign Phillips MP-70 monitor. To ensure synchronization between visual, vocal, and physiological data, we manually marked the start and end points of an event by simultaneously inserting that event to the Vital Sync\textsuperscript{TM} and using clapperboard with the video/audio stream. We made sure that our data collection setup does not cause any impediments for the regular clinical activities. Illustration of the equipment setup is shown in Fig. \ref{fig_equipment_setup}.

\subsection{Procedural Pain Dataset}
Procedural (Acute) pain data (visual, vocal, and vital signs) were collected from a total of 36 neonates (17 females) with a gestational age that ranges from 30 to 41 weeks. Data were collected prior to the painful procedures to get the baseline state, during scheduled painful procedures (e.g., heel lancing) and immediately after the procedures. Pain scores were provided for all periods (prior, during, and after) by trained nurses using the Neonatal Infant Pain Scale (NIPS) \cite{hudson2002validation}. This scale has three pain levels (no-pain, moderate, and severe) generated by combining scores of different pain indicators. Further description of the dataset is provided in \cite{zamzmi2019convolutional}. 

\subsection{Postoperative Pain Dataset}
Postoperative pain data (face, body, vocal, and vital signs) were collected from a total of 9 neonates (5 males) with a gestational age that ranges from 32 to 39 weeks. Three of the neonates are White, four Caucasian, and two Black. All the data were collected prior to a scheduled major surgery (e.g., omphalocele-repair) to get baseline state and up to three hours after the surgery to get postoperative pain state. Pain scores were provided by trained nurses for the baseline state as well as the postoperative state using N-PASS \cite{hummel2008clinical}. This scale has three pain levels (normal, moderate, and severe) generated by combining the scores of facial expression with the scores of other pain indicators. 

\begin{figure}[tb]
\centering
\includegraphics[width=.48\textwidth]{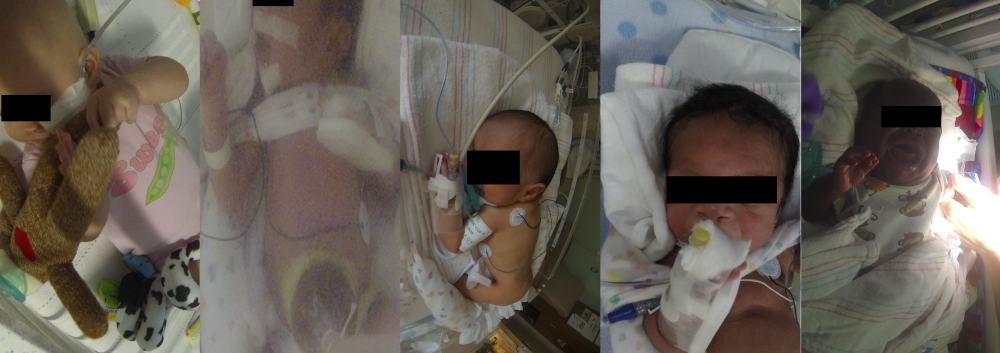}
\caption{Examples from our challenging and real-world postoperative dataset}
\label{fig_data_sample}
\end{figure}

Fig. \ref{fig_data_sample} shows examples from our postoperative dataset. The images were randomly selected and masked to ensure confidentiality. The challenges of the dataset can be easily observed from the images, which show face and body occlusion, different lighting conditions, noise, and cluttered background. 

\section{Methodology}
\begin{figure*}[t]
\centering
\includegraphics[width=.78\textwidth]{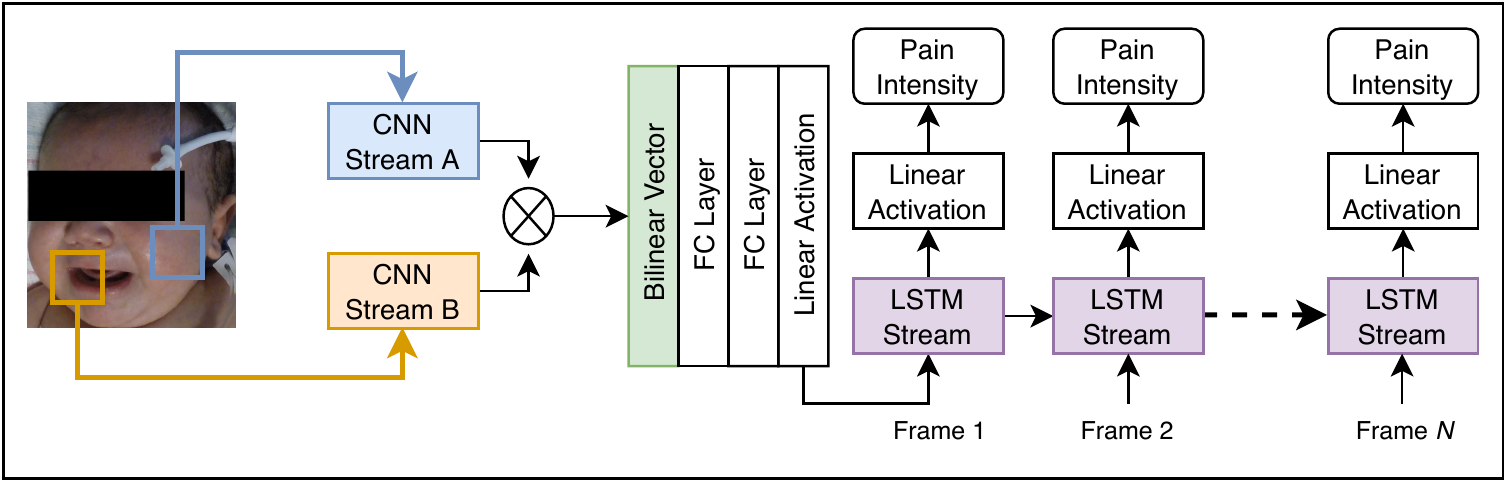}
\caption{Pipeline of the proposed framework for neonatal postoperative pain estimation. We used data of procedural and postoperative pain for training bilinear VGG16-based network; for LSTM training, we only used postoperative data. }
\label{fig_framework_pipline}
\end{figure*}

Our main aim is to automatically estimate the intensity of neonatal postoperative pain based on the provided N-PASS \cite{hummel2008clinical} scores. These scores are obtained from different indicators, including facial expression, crying sound, behavior states, extremities of tone, and vital signs. In this paper, we focus on estimating the intensity based on the analysis of facial expression only. We used both procedural and postoperative data to learn the spatial information of different levels of pain. We then used RNN to learn the unique pattern of postoperative pain over time. This temporal network (RNN) outputs the final intensity assessment of postoperative pain. We believe this approach can lead to better performance since our postoperative dataset is relatively small to train the network. Also, although the dynamic and intensity of postoperative pain are different from procedural pain, they both trigger a similar visual response or facial movements (spatial information). Therefore, we used both procedural and postoperative pain data for bilinear network training and only postoperative pain for LSTM training. The overall pipeline of the proposed approach is shown in Fig. \ref{fig_framework_pipline}.

\subsection{Pre-processing}
Pre-processing stage involves segmenting the recorded raw videos into pain and no-pain events. We found that the minimum length of a complete pain event in our dataset is $>$ 9 seconds. Therefore, we excluded all events with a length shorter than 9 seconds from further analysis. Because the number of the no-pain events is significantly larger than pain events, we balanced our dataset by reducing the number of events that belong to the majority class (no-pain); i.e., we under-sampled the no-pain events to balance it with the number of pain events. This sampling process generates 101 and 86 postoperative pain and no-pain events, respectively. 

\subsection{Key-Frame Extraction and Image Augmentation}
After segmenting all events, we extracted key-frames from these events using FFmpeg libraries. Since key-frames represent the position or timing of the movements within a given sequence or the points of any smooth transition, each sequence would generate a different number of key-frames. Instead of using varying lengths of key-frames, we used a fixed length, determined empirically, of size 32 key-frames. These frames are arranged in order and used as the representation for the video segment. We would like to note that we decided to use a fixed length of size 32 key-frames to obtain a vectorized representation with equal length, which allows to perform training or matrix operations in batch easily. To detect the face regions in all key-frames, we applied YOLO-based \cite{redmon2018yolov3} face detector, which was trained using the WIDER face dataset \cite{yang2016wider}, to each key-frame. The total number of final key-frames with detected face region, extracted from postoperative pain dataset, is 5984 ($101 \times 32$ + $86 \times 32$).

Finally, we performed image augmentation to enlarge our training set by randomly rotating all images up to 30 degrees, flipping them horizontally, and changing their brightness ($\pm 25\%$).

\subsection{Feature Extraction using Bilinear CNN Model}
We trained using both acute and postoperative data a bilinear CNN model \cite{lin2015bilinear} to extract features of different levels of pain. The billinear model allows to capture local pairwise feature interactions in a translationally invariant manner. For the bilinear CNN model, we used two VGG16 \cite{simonyan2014very} models, pre-trained with VGGFace2 \cite{cao2018vggface2} and Imagenet \cite{deng2009imagenet} dataset, as the backbones for the bilinear model. We then concatenated the features of both models (Stream A and B in Fig. \ref{fig_framework_pipline}) through the bilinear vector. Next, we added two fully connected layers of 64 units followed by a dense layer of 1 unit with linear activation. To prevent over-fitting, we used Dropout (0.5) layers between FC layers.

\subsection{Pain Intensity Estimation using LSTM Model}
As pain changes in a particular pattern over time, temporal information plays a critical role in estimating pain intensity. Hence, we used the RNN network, specifically LSTM \cite{hochreiter1997long}, to capture the temporal pattern and changes of postoperative pain over time. In particular, the facial features extracted by the bilinear CNN were passed to the LSTM network. As shown in Table \ref{tab_rnn_architecture}, we used two consecutive LSTM layers of 16 units, followed by two consecutive times distributed fully connected layers of 16 units. Finally, a dense layer of 1 unit with a linear activation function was used to continuously measure the pain intensity. We also used Dropout (0.3) layer after each time Distributed Dense Layer to avoid over-fitting.

\begin{table}[tb]
\centering
\caption{Details of LSTM Architecture} 
\begin{tabular}{|c|c|}
    \hline
    Layer Type & Configurations\\
    \hline
    \multirow{2}{*}{RNN} & LSTM 16, Activation = Tanh, \\
   & Recurrent Activation = Hard Sigmoid \\

    \hline
    \multirow{2}{*}{RNN} & LSTM 16, Activation = Tanh, \\
                         & Recurrent Activation = Hard Sigmoid\\
    \hline
    FC & Time Distributed Dense  16, Relu\\
    \hline
    FC & Time Distributed  Dense 16, Relu\\
    \hline
    FC & Time Distributed  Dense 1, Linear\\
    \hline
\end{tabular}
\label{tab_rnn_architecture}
\end{table}


\section{Experimental Results and Discussion}
\subsection{Dataset}
We report the results based on three datasets: our acute dataset, COPE acute dataset \cite{brahnam2007introduction}, and our postoperative dataset. Before proceeding, we note that the intensity of facial expression ranges from 0 (no-pain) to 1 (pain) in case of acute pain (our dataset and COPE). In the case of postoperative pain, the intensity of facial expression is scored as 0 (no-pain), 1 (moderate pain), and 2 (strong pain). Because the number of images with score 2 is very small, we combined images of score 1 and score 2 into one pain class. This also allows to generate the same range of facial pain levels ($0-1$) of the acute pain.

In addition to the key-frames extracted from our dataset, we used static images of the COPE dataset \cite{brahnam2007introduction} as a second acute pain dataset for training the bilinear CNN. This dataset consists of 204 static images of 26 babies (13 females) with a gestational age that ranges from 18 hours to 3 days. The images were collected during four different stimuli: crib change, air stimulus, friction, and pain (heel lancing). In this work, we divided these images into pain images (heel lancing) and no-pain images (crib change, air stimulus, friction). We rotated and aligned COPE images to obtain the frontal view of the face.

Finally, the proposed LSTM model is used to estimate the final pain intensity of the postoperative pain. In N-PASS \cite{hummel2008clinical}, the total pain score ranges from -10 to +10. Table \ref{tab_postoperative} shows the distribution of pain intensity levels in our postoperative dataset. Note that our dataset has instances for only 8 intensity levels ($0-7$).

\begin{table}[tb]
\centering
\caption{Pain intensity ($0-7$) distribution of the neonatal postoperative pain after pre-processing}
\begin{tabular}{|c|c|}
\hline
    Pain intensity  & Number of images  \\ \hline\hline
    0 & 1728     \\ \hline
    1 & 160     \\ \hline
    2 & 512     \\ \hline
    3 & 352     \\ \hline
    4 & 928     \\ \hline
    5 & 960     \\ \hline
    6 & 352     \\ \hline
    7 & 992     \\ \hline \hline
    Total  & 5984     \\ \hline
\end{tabular}
\label{tab_postoperative}
\end{table}

\subsection{Training and Evaluation Protocol}
In our experiments, we resized all key-frames to the standard input size ($224 \times 224$) for most CNN models. For the bilinear VGG16-based \cite{lin2015bilinear} model, we fine-tuned the model from the last convolutional blocks, which means after the fourth pooling layers of the network. Similar to this \cite{wang2017regularizing}, both models were trained using a regression function instead of classification which is more suitable to the nature of the problem. We used mean square error (MSE) to measure the loss. We trained the bilinear CNN model using gradient Adam optimizer with a learning rate of 0.0001 and a batch size of 16. In the case of LSTM, features were extracted by the bilinear model and fed into the LSTM network. We used MSE to measure the loss and Adam optimizer with a learning rate of 0.0001. 

For both CNN and LSTM training, we used leave-one-subject-out evaluation protocol for postoperative pain. While training the CNN models with acute datasets, we randomly split the dataset to 80\% (training and validation) and 20\% (testing). All the models were trained using 150 epochs and early stopping was used to avoid over-fitting.

\subsection{Results Analysis}
We trained CNN models using a regression function to estimate the pain intensity of the expression. Table \ref{tab_cnn} shows the performance of (a) VGG16 alone and (b) bilinear VGG16 reported using the mean square error (MSE) and mean absolute error (MAE). VGG16 CNN (a) \cite{simonyan2014very} alone was used to model facial pain intensities. Similar to \cite{zamzmi2019convolutional}, we used a shallow network for VGG16 which is shown to be optimum for relatively small datasets. We replaced the last 3 fully connected layers by two dense layers of 512 units, followed by a dense layer of 1 unit with a linear activation function. To prevent over-fitting, we used Dropout (0.5) layers between Fully Connected (FC) layers. We fine-tuned the fully connected layers of VGG16 model (a) with images from COPE dataset, our acute, and postoperative pain dataset. This baseline model (a) was trained using a regression function and gradient Adam optimizer with a learning rate of 0.0001 and a batch size of 16. The architecture and training parameters of the bilinear VGG6-based CNN (b) are presented in Fig. \ref{fig_framework_pipline} and Section V.B. 

As shown in the Table, bilinear VGG16 performs better than VGG16 and fine-tuning the models with our acute pain dataset achieves better performance than fine-tuning with COPE \cite{brahnam2007introduction} dataset. The Table also shows that the performance of acute pain is higher than postoperative in all cases. This can be attributed to the small number of subjects and the low variations of pain intensities in the postoperative pain data as compared to the acute, which has a larger number of subjects with higher variations. In most cases, the bilinear VGG16 achieves better performance than the regular VGG16, which suggests that pain intensity estimation can be considered a fine-grained classification problem. To measure the temporal changes of postoperative pain, we used the best two feature extractors (2nd and 6th in Table \ref{tab_cnn}) to train the LSTM. The performance of pain intensity assessment using the best CNN models with LSTM is presented in Table \ref{tab_rnn}. The results suggest that bilinear CNN can better represent the facial pain intensity feature than the general CNN model. These results are encouraging and suggest the efficiency of using a combination of bilinear CNN with LSTM for assessing postoperative pain intensity. 

\begin{table}[tb]

\centering
\caption{Neonatal post-op assessment (facial pain intensity [0-1])}
\begin{tabular}{|l|c|c|c|c|}
\hline
    Approach       & Pretrain               & Retrain    & MSE      & MAE  \\ \hline\hline
    VGG16          & VGGFace2               & COPE       & 0.4170   & 0.5412\\ \hline
    VGG16          & VGGFace2               & Acute      & 0.1979   & 0.4035\\ \hline
    VGG16          & VGGFace2               & Post-Op    & 0.3606   & 0.5155\\ \hline   
    VGG16          & Acute                  & Post-Op    & 0.3716   & 0.5211\\ \hline
    Bilinear VGG16 & VGGFace2, ImageNet     & COPE       & 0.4272   & 0.5208\\ \hline
    Bilinear VGG16 & VGGFace2, ImageNet     & Acute      & 0.1917   & 0.3458\\ \hline
    Bilinear VGG16 & VGGFace2, ImageNet     & Post-Op    & 0.2955   & 0.4575\\ \hline
    Bilinear VGG16 & Acute                  & Post-Op    & 0.2695   & 0.4173\\ \hline
\end{tabular}
\label{tab_cnn}
\end{table}

\begin{table}[tb]
\centering
\caption{Neonatal post-op assessment (pain intensity [0-7]) using LSTM}
\begin{tabular}{|l|c|c|}
\hline
    Approach                 & MSE        & MAE  \\ \hline\hline
    VGG16 + LSTM             & 4.8612     & 1.7274 \\ \hline
    Bilinear VGG16 + LSTM    & 3.999      & 1.5565  \\ \hline
\end{tabular}
\label{tab_rnn}
\end{table}


\section{Conclusion and Future Work}
This paper proposes a fully automated deep learning-based system that aims to mitigate the limitations of the current assessment practice by providing standardized and continuous monitoring of neonatal postoperative pain. A bilinear CNN model followed by LSTM architecture is proposed to estimate the neonatal postoperative pain intensity. The experimental results are encouraging and prove the feasibility of using the proposed framework for assessing neonatal postoperative pain. Future work will focus on evaluating the proposed framework on a larger postoperative pain dataset and incorporating other pain modalities.



\bibliographystyle{ieeetr}
\bibliography{bibliography}

\end{document}